\documentclass[10pt,twocolumn,letterpaper]{article}

\usepackage[pagenumbers]{wacv} 

\usepackage[breaklinks,colorlinks]{hyperref}

\usepackage{graphicx}
\usepackage{amsmath}
\usepackage{amssymb}
\usepackage{booktabs}

\usepackage{comment}
\usepackage{multirow} 
\usepackage{float}

\newcommand{\name}{AGA }
\newcommand{\nameNoSpace}{AGA}

\usepackage[capitalize]{cleveref}
\crefname{section}{Sec.}{Secs.}
\Crefname{section}{Section}{Sections}
\Crefname{table}{Table}{Tables}
\crefname{table}{Tab.}{Tabs.}

\def\wacvPaperID{*****} 
\def\confName{WACV}
\def\confYear{2025}

\begin{document}

\title{Data Augmentation for Image Classification using
Generative AI}

\author{
Fazle Rahat\\
University of Central Florida\\
Orlando, FL\\
{\tt\small fazle.rahat@ucf.edu}
\and
M Shifat Hossain\\
University of Central Florida\\
Orlando, FL\\
{\tt\small mshifat.hossain@ucf.edu}
\and
Md Rubel Ahmed\\
University of Central Florida\\
Orlando, FL\\
{\tt\small mdrubel.ahmed@ucf.edu}
\and
Sumit Kumar Jha\\
Florida International University\\
Miami, FL\\
{\tt\small jha@cs.fiu.edu}
\and
Rickard Ewetz\\
University of Florida\\
Gainesville, FL\\
{\tt\small rewetz@ufl.edu}
}

\maketitle

\begin{abstract}
Scaling laws dictate that the performance of AI models is proportional to the amount of available data. Data augmentation is a promising solution to 
expanding the dataset size. Traditional approaches focused on augmentation using rotation, translation, and resizing. Recent approaches use generative AI models to improve dataset diversity.
However, the generative methods struggle with issues such as subject corruption and the introduction of irrelevant artifacts. 
In this paper, we propose the \underline{A}utomated \underline{G}enerative Data \underline{A}ugmentation (\nameNoSpace).  
The framework combines the utility of large language models (LLMs), diffusion models, and segmentation models to augment data. \name preserves  foreground authenticity while ensuring background diversity. Specific contributions include: i) segment and superclass based object extraction, ii) prompt diversity with combinatorial complexity using prompt decomposition, and iii) affine subject manipulation. 
We evaluate \name against state-of-the-art (SOTA) techniques on three representative datasets, ImageNet, CUB and iWildCam. The experimental evaluation demonstrates an accuracy improvement of $15.6\%$ and $23.5\%$ for in and out-of-distribution data compared to baseline models respectively. There is also  64.3\% improvement in SIC score compared to the baselines. 
\end{abstract}
\vspace{-12pt}
\section{Introduction}
\label{sec:intro}
Deep learning models often struggle with domain adaptation when exposed to new conditions, such as attacks~\cite{walmer2022dual,michel2022survey,lyu2024task}, changes in weather~\cite{ramanathan2017adversarial,ozdag2019susceptibility}, and geographic locations~\cite{Huang_2018_ECCV,jha2022responsible}. This issue is particularly evident in applications like rare bird or animal species identification, where insufficient training data can hinder the model's ability to generalize effectively~\cite{shorten2019survey}. Adding more training data from diverse domains can help alleviate this issue; however, collecting high-quality and relevant data is inherently costly~\cite{roh2019survey}.

Significant research efforts have been dedicated to traditional data augmentation approaches based on geometric modifications, including cropping, translations, and rotations~\cite{shorten2019survey}. The limitations of these techniques is that the subject features may be altered and the limited image diversity. On the other hand, the recent  advancements within generative AI is providing new opportunities for data augmentation~\cite{shorten2019survey} using large language models (LLMs)~\cite{devlin2018bert},
vision-language models (VLMs)~\cite{li2022blip,su2022language}, image synthesis models~\cite{podell2023sdxl,NEURIPS2021_9219adc5}. In particular, the ability to synthesize photo realistic images from natural language~\cite{NEURIPS2021_49ad23d1, NEURIPS2020_4c5bcfec, sohl2015deep}. 
These models demonstrate exceptional performance on various tasks such as text-to-image generation~\cite{rombach2022high, azizi2023synthetic}, image-to-image modification~\cite{meng2021sdedit, dunlap2024diversify}, and image inpainting \cite{lugmayr2022repaint}. 
Recent work shows that large-scale diffusion models can be fine-tuned to generate augmented images for improving recognition tasks~\cite{azizi2023synthetic}. 
While fine-tuning image generation models for data augmentation is effective, their complexity and the need for replication across diverse datasets often make it impractical~\cite{yu2023diversify}. Methods for augmenting visually realistic images using text-guided techniques without model fine-tuning are proposed in~\cite{bansal2023leaving, yu2023diversify, dunlap2024diversify}. However, our case study indicates that diffusion models struggle to augment fruitful training data from text prompts alone, often deviating from the intended subjects in the generated images.

\begin{figure*}[h]
\centering
\includegraphics[width=0.99\textwidth]{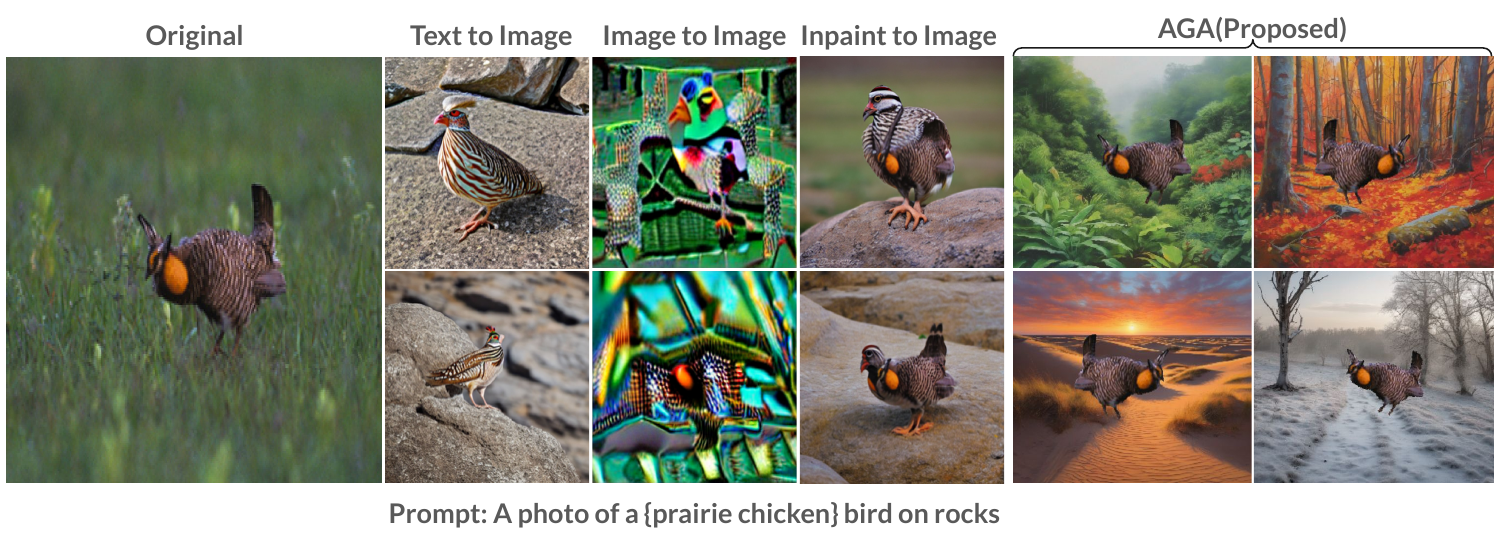}
\vspace{-12pt}
\caption{Example augmentation using text-to-image, image-to-image, inpainting, and our approach on ImageNet10. Images generated by text-to-image and image-to-image significantly lose foreground information. Inpainting provides comparatively better results but corrupts the foreground with unnecessary modifications. \name is able to produce diverse background images while keeping the foreground information grounded with original images.}
\label{fig:casestudy}
\vspace{-10pt}
\end{figure*}

 In this paper, we propose the Automated Generative Data Augmentation framework called \name to augment the training dataset to enhance fine-grained classification performance.
Our method aims to alter the subjects minimally while introducing variability in the backgrounds during the augmentation process. AGA uses image segmentation to isolate subjects, a pre-trained LLM for varied background captions, Stable Diffusion for diverse background creation, and integrates subjects seamlessly with backgrounds. 
Automatic background image generation faces two main challenges. The first is creating diverse backgrounds without corrupting the foreground, a problem often overlooked by existing methods effectively addressed by the subject isolation technique of \name.
 The other challenge is creating the right LLM prompt automatically. \name solves this by including a prompt generation engine equipped with hierarchical instruction, spatial and temporal modality fixers. This engine automatically produces a diverse set of text prompts for the LLM while using a small library of sample instructions, which ultimately ensures the diversity in the generated backgrounds. 
This paper makes the following key contributions:
\begin{itemize}
    \vspace{-1mm}
    \item We introduce \name, an innovative framework for data augmentation that focuses on diversifying backgrounds while preserving the subject of interest with various affine transformations, leading to robust and explainable classifiers.
    \vspace{-2mm}
    \item \name leverages large language and vision models to automatically create diverse training images, eliminating the need for manual user input or fine-tuning.  
    
    \vspace{-2mm}
    \item \name improves the accuracy of fine-grained classification from $78.4\%$ to $93.6$\% on a reduced version of the ImageNet dataset.
    
\end{itemize}

The rest of the paper is organized as follows: Section~\ref{sec:literature} discusses related  work, Section~\ref{sec:case_study} motivates our work with a case study, Section~\ref{sec:method} presents our methodology. The experimental results are discussed in Section~\ref{sec:exp} and finally, Section~\ref{sec:conclusion} concludes the paper with potential future work.

\section{Related Works}
\label{sec:literature}

Image augmentation is a pivotal method for improving the performance and generalization ability of deep learning models. Early works often resort to geometric transformations such as flipping, cropping, and rotation, color space transformations, kernel filters~\cite{shorten2019survey}. Beyond simple manipulation, advanced techniques like Mixup~\cite{zhang2017mixup} and CutMix~\cite{Yun_2019_ICCV} introduce advanced techniques such as mixing images to create new training examples and encourage the model to learn more robust representations. Additionally, automated augmentation methods such as RandAugment~\cite{Cubuk_2020_CVPR_Workshops} randomly select and apply a sequence of transformations with varying magnitudes, eliminating the need for manual tuning of augmentation hyperparameters. However, these  techniques often generate  images which are not only visually unnatural~\cite{dunlap2024diversify} but also loses subject information.

With the advent of generative AI models, particularly diffusion models, image augmentation has witnessed a paradigm shift and these models are widely adopted in image generation~\cite{azizi2023synthetic, NEURIPS2021_49ad23d1, rombach2022high, dunlap2024diversify, yu2023diversify}. Large-scale image-text datasets and models like CLIP~\cite{radford2021learning} have enabled SOTA diffusion models to perform versatile tasks such as text-to-image generation, image-to-image transformation, and inpainting through text-guided prompts. Several studies have investigated 
 how to enhance the classification accuracy using  
 synthetic images generated by diffusion models~\cite{he2022synthetic, lin2023explore}. 
 One study showed that it is possible to train a classifier for ImageNet solely using synthetic Data, leading to a performance improvement when applied to real-world tasks~\cite{sariyildiz2023fake}. While another investigation demonstrates the effectiveness of fine-tuning Imagen~\cite{saharia2022photorealistic} for data augmentation on ImageNet~\cite{azizi2023synthetic}. These fine-tuning based approaches face practical challenges due to complexity, cost, and dataset-specific requirements. Recent works utilize off-the-shelf diffusion models  to diversify vision datasets  without the need for fine-tuning~\cite{yu2023diversify, dunlap2024diversify}.

Methods for attribution analysis of neural networks that seek to explain an AI decision~\cite{jha2019explaining,jha2021smoother, jha2022shaping, jha2022explainit, walker2024integrated, jha2023neural, jha2023neural, kumari2024towards} can benefit from data sets where the background has been substantially altered. Methods from assurance analysis of AI systems~\cite{jha2019attribution,langmead2007predicting,jha2018detecting,jha2020model,jha2021protein} can greatly benefit from analysis of scenarios where the background has been replaced with diverse contexts. Methods for out-of-context object detection~\cite{kaur2021detecting,magesh2023principled,kaur2023predicting} and analysis of visual-language models~\cite{mangal2024concept} can readily benefit from the identification and diversification of image backgrounds. All of these methods can benefit from our approach of data augmentation using background diversification.

Recent works create synthetic images using  either text-to-image~\cite{rombach2022high} or image-to-image~\cite{meng2021sdedit} methods, with text-guided image generation. Image-guided inpainting~\cite{lugmayr2022repaint} also utilizes image modification to introduce diversity in the image data. However, these techniques significantly distort required subject information. To solve this issue and generate synthetic images without losing subject information, we propose \name. This is an automatic segmentation-guided technique that utilizes recent object detection and segmentation models~\cite{kirillov2023segment, liu2023grounding} to augment data. \name generates effective synthetic images while keeping foregrounds grounded with the original images.

\begin{figure*}[h]
\centering
\includegraphics[width=0.99\textwidth]{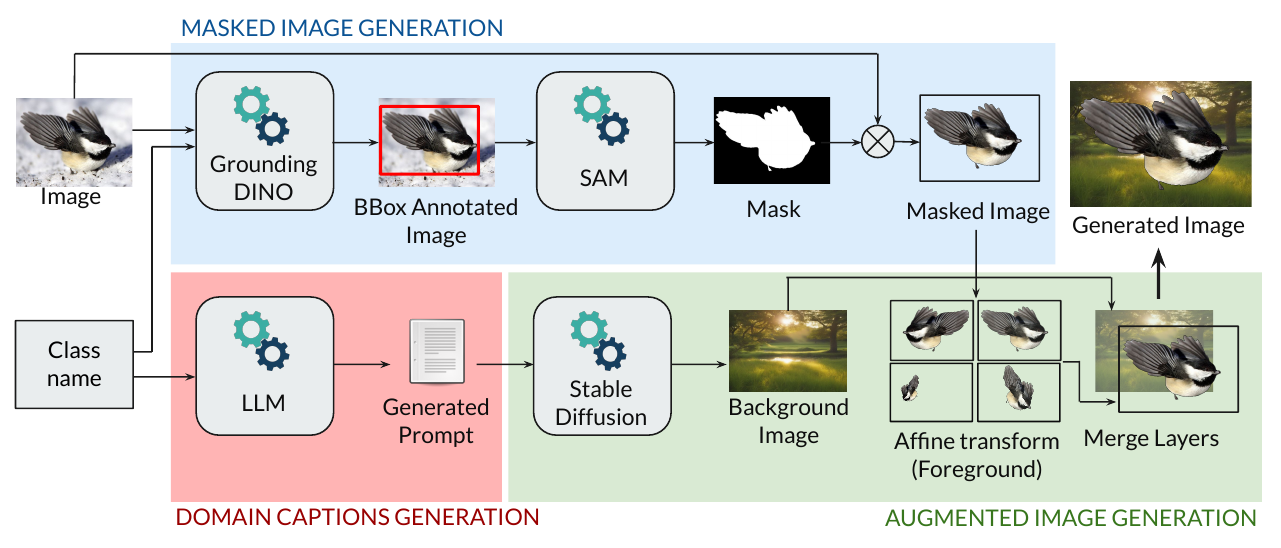}
\caption{The methodology of the \name framework. The inputs are an image and originals class name, while the outputs are corresponding augmented images. Subject isolation from input is performed by masked image generation. The domain captions generation engine generates diverse background prompts, which are utilized by stable diffusion to generate background images. Finally, these background images and isolated subjects are combined to generate augmented images.}
\label{fig:system_overview}
\vspace{-10pt}
\end{figure*}

\section{A Motivating Case Study}
\label{sec:case_study}

Text-to-image, image-to-image, and inpainting are three key image augmentation techniques extensively utilized in recent image augmentation works. We conducted a case study of these methods for several datasets like ImageNet~\cite{deng2009imagenet}, CUB~\cite{wah2011caltech}, and iWildCam~\cite{koh2021wilds}, to understand their advantages and shortcomings. We discuss our observations using a representative image of a bird from the CUB data set, as illustrated in Figure~\ref{fig:casestudy}.

\noindent{\bf Text-to-image:} It can be observed that text-to-image, while capable of generating a high diversity of images, often produces samples in which the subject is so drastically altered that even human observers struggle to identify it. We find in the figure that the identifying mark of the bird, the red ring around the neck, is missing in the augmented images, which would translate into failures for downstream tasks.

\noindent{\bf Image-to-image:} This type of augmentation frequently results in a significant loss of subject detail, akin to the image-to-text method. We see that the bird is very hard to spot in the augmented image, which might make the downstream object detection task tougher. The prairie chicken in the example appears to have been transformed into a parrot.

\noindent{\bf Inpainting:} This method operates by modifying the image within a masked area based on a text prompt, yet this method can inadvertently corrupt the subject's appearance. We see that the orange/red identifying ring misplaced in one of the augmented images, thus this method can corrupt the subject. 

 In contrast, our proposed approach does not add any artifacts to the subject image. It successfully generates augmented images with diverse backgrounds while preserving the authenticity of the subject as seen in the augmented images by \name. This form of data augmentation has the potential to translate into better performance in terms of fine-grained classification, generalizability, and explainability. Our proposed approach is presented in the next section.
 \vspace{-4pt}

\section{Automated Generative Data Augmentation}
\label{sec:method}

In this section, we present the methodology of the \name framework. The input to the framework is an image and the corresponding class name. The output is an augmented image based on the provided inputs. The framework augments an image in three main  steps: i) subject isolation through \emph{masked image generation}, ii) the generation of \emph{domain-specific captions} for diverse backgrounds, and iii) augmented image editing for combining the foreground and background. An overview of the \name framework is shown in Figure~\ref{fig:system_overview}.

\subsection{Masked Image Generation}
\label{sec:masked-images}

This step deals with isolating the subject of an input image from its background. In general, such subject masks are not readily available beside the image and class name. Therefore, dense mask estimation models can be used to correctly generate pixel-level masks for subjects using the image and text (class name) only. 

 \name includes Segment Anything Model (SAM)~\cite{kirillov2023segment}, one of the SOTA image segmentation tool, for this purpose. SAM is capable of segmenting the subject from an image based on some guiding inputs such as single or multiple point locations on an object, or the object's bounding box, to create precise segmentation masks. As the training dataset usually does not include the point locations or bounding box for the subject,  object detection models can be utilized to generate the boxes in this regard. Bounding boxes provide approximate spatial locations of objects of our interest in the image. 

\begin{figure}[h]
\centering
\includegraphics[width=0.99\linewidth]{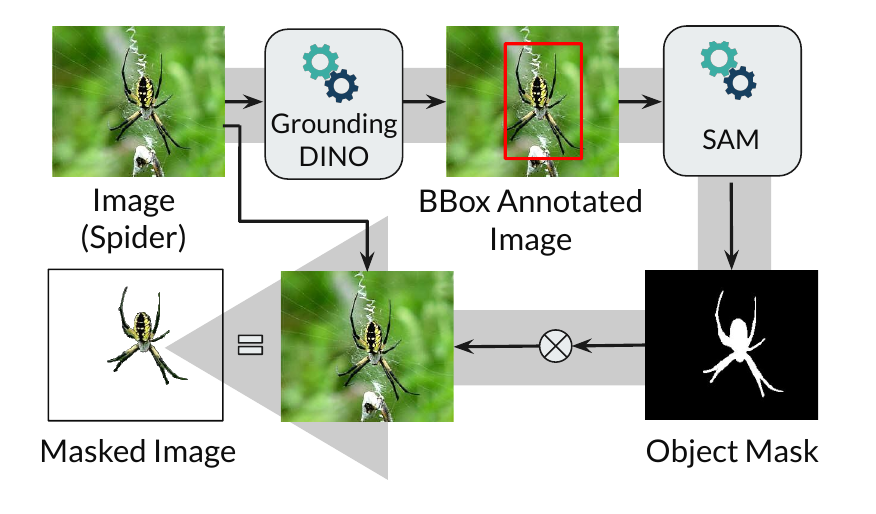}
\vspace{-12pt}
\caption{Masked image generation process diagram}
\vspace{-7pt}
\label{fig:mask_image}
\end{figure}

There are several SOTA object detection models available, like YOLO~\cite{Redmon_2016_CVPR}, GoundingDINO~\cite{liu2023grounding}. In \name workflow, GroundingDINO model is used to generate bounding box due to its superior performance. Empirical analysis shows that for fine-grained text prompts, GroundingDINO often fails to provide optimal bounding box results. For instance, when attempting to locate the bounding box for a specific bird class such as \texttt{water~ouzel}, the hierarchical naming of the class text \texttt{bird} proves more effective than the fine-grained class name. Therefore, \name utilizes superclasses as text prompts to provide clearer input instructions to the GroundingDINO model for object bounding box creation.

The details of proposed mask generation process is shown in Figure~\ref{fig:mask_image}. GroundingDINO generates the bounding box, indicated in red, for the subject of interest, which in this case is the spider. This bounding box is then feed to the SAM to guide itself to produce the segmentation mask. Once the mask is obtained, the masked image of the subject is created by combining the original image with the segmentation mask.

\subsection{Domain Caption Generation} 
\label{sec:generate-prompt}

In the \name pipeline, the generation of domain captions is a crucial task, as it directly influences the diversity of the background images produced. These captions are automatically generated through a two-step process using a prompt generation engine. Initially, the engine samples from three predefined sets: the instruction set ($Ins$), the background set ($Bgr$), and the temporal modality set ($Temp$) as the prompt fixers. The instruction set ensures the prompt begins with an appropriate command, the background set introduces spatial diversity, and the temporal set enriches the prompt with times of the day and seasons.

\begin{figure}[!ht]
\centering
\vspace{-3pt}
\includegraphics[width=0.99\linewidth]{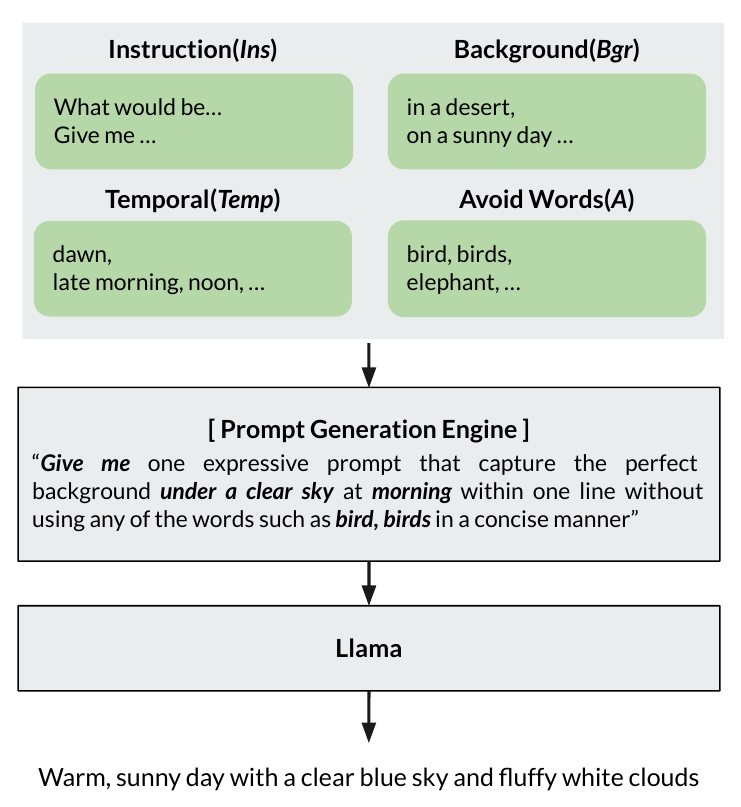}
\caption{Prompt generation for background diversity.}
\vspace{-15pt}
\label{fig:prompt_eng}
\end{figure}

A SOTA LLM, Llama, is employed to transform these engineered prompts into detailed captions that guide the vision diffusion model in generating the background images. Furthermore, a list of words to avoid is incorporated to refine the output, ensuring the prompts remain focused and relevant. The words to avoid include the class names or subject of the image dataset to be augmented, as those subject might corrupt the background prompt.
This structured approach ensures each dataset receives tailored prompts, enhancing the resulting image diversity, which is described in Figure~\ref{fig:prompt_eng}. Moreover, each part of the prompt results in a combinatorial increase in diversity. This reduces the number of prompt samples that are required to be provided for each category of instructions.

\subsection{Augmented Image Generation}

\label{sec:edit-images}

In this step, utilizing the masked image obtained from Section~\ref{sec:masked-images} and the background caption prompt from Section~\ref{sec:generate-prompt}, \name generates a new image with an altered background. The caption prompt serves as the input for a large vision model, which is responsible for creating the background image. Among several text-to-image generation models available, such as DALL-E~\cite{pmlr-v139-ramesh21a}, Imagen~\cite{saharia2022photorealistic}, and Stable Diffusion~\cite{rombach2022high}, \name employs the Stable Diffusion model for this purpose.
Once we have both the masked image and the background image produced by the vision model, \name proceeds to create the new augmented image. The merging technique used ensures that the background image fills all areas with the masked image, except where the subject is located. This method allows the subject to remain prominent against the newly generated background.

\begin{figure}[htbp]
\centering
\includegraphics[width=0.49\textwidth]{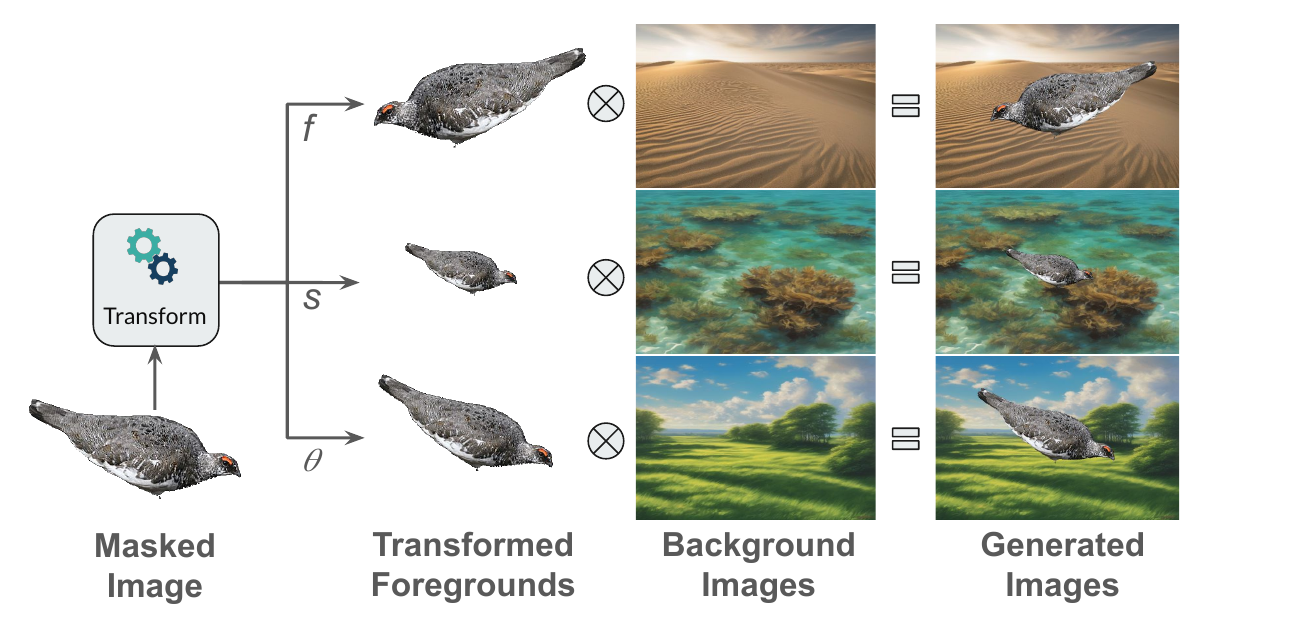}
\caption{Merging image mask with the generated backgrounds while utilizing affine transformations.}
\label{fig:edit_image}
\end{figure}

Additionally, to enhance diversity without altering the semantic content of the image, \name applies traditional affine transformations to the masked image prior to merging. These transformations include flipping (\textit{f}), rotating ({\it $\theta$}), and scaling (\textit{s}) the subject. Figure~\ref{fig:edit_image} illustrates these image editing processes and the respective transformations, showcasing how they contribute to the diversity of the final augmented image while preserving its original meaning.

\section{Experimental Evaluation}
\label{sec:exp}

We implement the AGA framework in Python, utilizing open source APIs for machine learning models. The implementation runs on a machine equipped with an NVIDIA A100 graphics card. The following sections provide detailed descriptions of the dataset preparation, evaluation setup, and key findings.
\paragraph{Setup.} 

We created a subset of the ImageNet~\cite{deng2009imagenet}  dataset, named ImageNet10, by randomly selecting $10$ classes. This subset comprises $13,046$ and $500$ training and validation images across the following classes. We refer to this trainset as the original dataset and generate synthetic images from it using the AGA methodology. In subsequent discussions, models described as trained with augmented data refer to those trained using both the original and augmented datasets. Beside this ImageNet10 dataset, we utilize iWildCam~\cite{koh2021wilds} dataset, which contains a large collection of global camera trap images. %
Similarly, we extend our experiments to the CUB~\cite{wah2011caltech} dataset, a fine-grained classification set of $200$ bird species from Flickr. We maintain the same data distribution ratio as in the previous work`\cite{dunlap2024diversify} for train and test set to ensure a fair comparison. Detailed dataset descriptions are included in the supplementary materials.

We evaluate the \name method across two main categories. First, we conduct an in-distribution evaluation of our pipeline using the ImageNet validation dataset for each model. Second, we assess the robustness of our augmentation method by evaluating it on out-of-distribution ImageNet data. For this we use the ImageNet variations: ImageNet-Sketch~\cite{wang2019learning} and ImageNet-V2~\cite{recht2019imagenet} where ImageNet-Sketch is the sketch version and ImageNet-V2 is the reproduced version of ImageNet respectively. The CUB and iWildCam datasets are used to conduct a comparison study with the previous work~\cite{dunlap2024diversify}. We consider two types of models for our experiments: those trained with original image data and those trained with augmented image data. For comparison, we maintain baseline hyperparameters while augmenting the original training data with augmented data at various scales.

We also compare with other augmentation techniques from recent times: (1) MixUp~\cite{zhang2017mixup}, a data augmentation technique improves deep learning model generalization by creating virtual training examples through convex combinations of original data points and labels, to enhance model robustness and performance on unseen data. (2) CutMix~\cite{Yun_2019_ICCV}, an approach that creates mixed samples by randomly slicing and combining patches from multiple training images to capture finer and more distinct features across localized regions. (3) RandAugment~\cite{Cubuk_2020_CVPR_Workshops}, simplifies data augmentation by reducing the search space for augmentation strategies, automating the selection of operations and magnitudes to enhance model performance and generalization while minimizing computational overhead. (4) ALIA~\cite{dunlap2024diversify} analyzes training images to identify diverse background captions, then uses this information to create variations of the images with different backgrounds and contexts. After that, it removes any images that are low-quality or corrupt the original information.
In addition to these experiments,  we include the results of explainability enhancements for machine learning models using AGA augmented data.

\begin{figure}[t]
\centering
\includegraphics[width=0.5\textwidth]{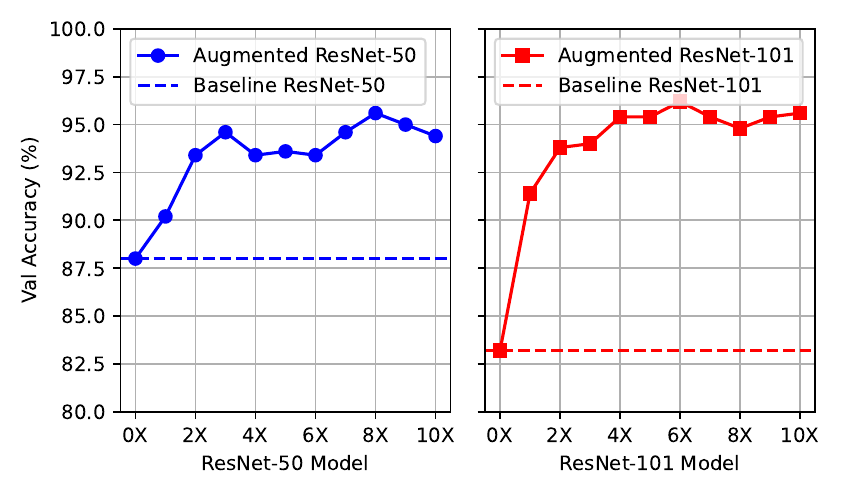}
\caption{The figure shows the top-1 accuracy with respect to the degree of data augmentation (AGA) for the ResNet-50 and ResNet-101 models on ImageNet10. It can be observed that the accuracy rapidly improves in the beginning while showing an overall upward trend until 10 times augmentation.} 
\vspace{-15pt}
\label{fig:scaling_up}
\end{figure}

\paragraph{Implementation.} 
We employ ResNet variants {18, 50, 101, 152} as the classification models for training. We train these CNN models from scratch using PyTorch's standard training script~\cite{torchvision2016} which includes PyTorch's default hyperparameter set~\cite{pytorchTrainStateOfTheArt}. \name utilize a Llama-2-13B-GPTQ from Hugging-Face~\cite{von-platen-etal-2022-diffusers} to create background image caption prompts. These prompts are generated for each image using our prompt engineering method outlined in Section~\ref{sec:generate-prompt}. Background images are then generated using the Stable Diffusion XL~\cite{podell2023sdxl} text-to-image model from Hugging-Face, with default hyperparameters.
The prompt generation engine operates with three distinct modality sets: an instruction, spatial, and temporal modality set size of $3$, $18$, and $13$ respectively. Supplementary materials include additional training and hyperparameter descriptions.

The remainder of this section is organized, as follows: We first evaluate the effectiveness of \name on fine-grained image classification in Section 5.1. Next, the generalizability is evaluated in Section 5.2. Lastly, the impact on the explainability is evaluated in Section 5.3. 
Supplementary materials contain additional evaluation results.

\begin{figure}[t]
\centering
\includegraphics[width=0.99\linewidth]{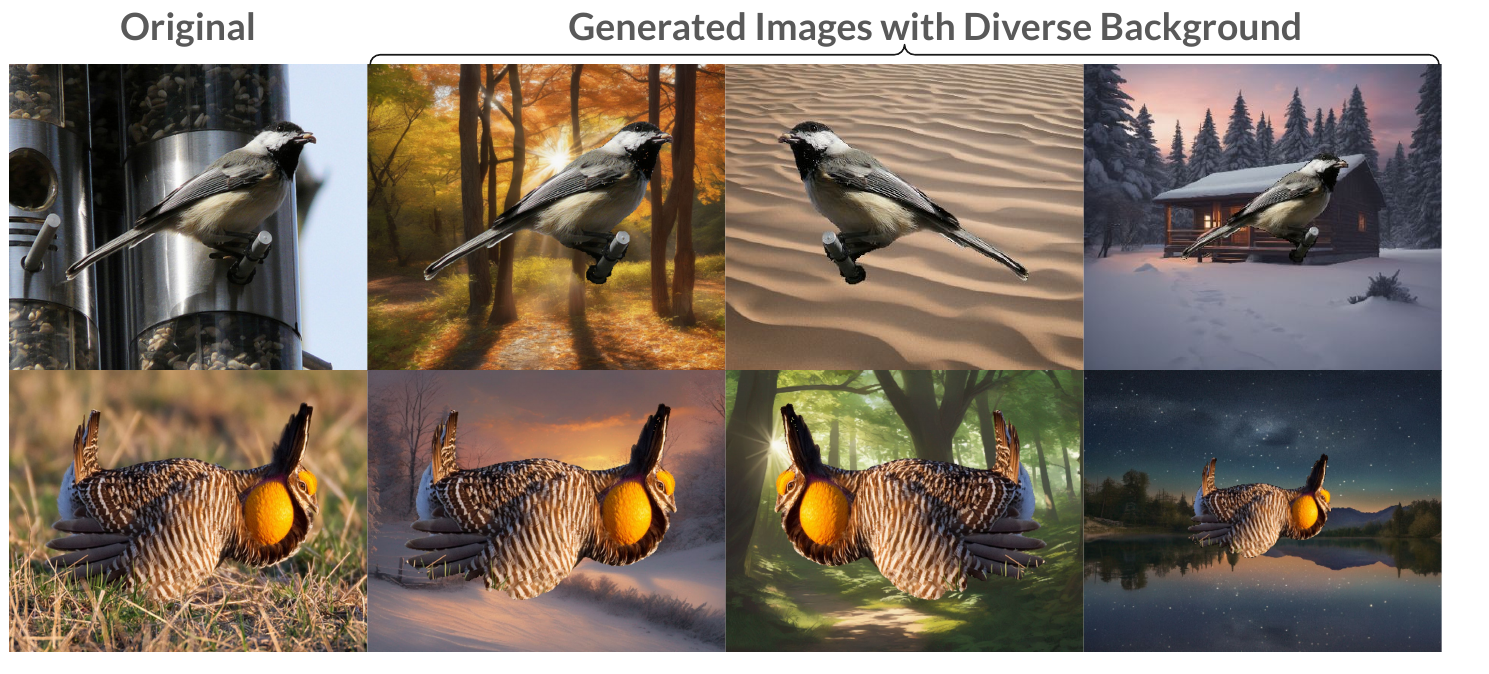}
\caption{The figure displays the original image samples from ImageNet10 and the generated images using \name. The generated images effectively preserve the authenticity of the subjects while exhibiting diverse backgrounds. Additional synthetic images are shown in supplementary element.}
\vspace{-15pt}
\label{fig:imagenet10_results}
\end{figure}

\begin{table*}[htbp]
\centering
\caption{{\bf Top-1 Accuracy on ImageNet10 and its OOD variations.} Our approach involves training models with both real ImageNet10 training images and synthetic images generated through our pipeline. We perform an in-distribution evaluation on ImageNet10-val and out-of-distribution evaluation on ImageNet-V2 and ImageNet-Sketch datasets. We directly applied the trained models to these out-of-distribution evaluation datasets without further fine-tuning. The improvement ($\Delta$) over baseline models trained with only real images.}
\renewcommand{\arraystretch}{1.1} %
\small
\begin{tabular}{l|lll|lll|lll}
    \hline
     \multirow{3}{*}{\textbf{Model}} & \multicolumn{3}{|c}{\textbf{In-Distribution}} & \multicolumn{6}{|c}{\textbf{Out-of-Distribution}} \\
    \cline{2-10}
    \multicolumn{1}{c}{\multirow{2}{*}{{\underline{\hspace{0cm}}}}} & \multicolumn{3}{|c|}{\textbf{ImageNet10-Val}} & \multicolumn{3}{c}{\textbf{ImageNet-V2}} & \multicolumn{3}{|c}{\textbf{ImageNet-S}} \\ \cline{2-10} 
    \multicolumn{1}{c|}{} & Baseline & \name & $\Delta$  & Baseline & \name & $\Delta$ & Baseline & \name & $\Delta$ \\ 
    \hline 
    ResNet-18 & 88.8\% & 93.4\% & {4.6} & 78.43\% & 89.21\% & 10.78 & 33.46\% & 47.75\% & 14.29 \\
    ResNet-50 & 86.8\% & 94.6\% & 7.8 & 81.37\% & 85.29\% & 3.92 & 27.78\% & 50.09\% & {\textbf{22.31}} \\
    ResNet-101 & 78.4\% & 93.6\% & {\textbf{15.6}} & 65.69\% & 89.22\% & {\textbf{23.53}} & 28.77\% & 46.77\% & 18.0 \\
    ResNet-152 & 81.6\% & 93.8\% & 12.2 & 72.55\% & 88.24\% & 15.69 & 27.98\% & 46.57\% & 18.59 \\ \hline
\end{tabular}
\label{tab:in_out_comparison}
\end{table*}

\subsection{Fine-grained Image Classification}

\textbf{Accuracy vs. Degree of Data Augmentation:} We first evaluate the improvement in classification accuracy with respect to the amount of data augmentation in Figure~\ref{fig:scaling_up}. The figure shows  the classification accuracy of a ResNet-50 and a ResNet-101 model on the ImageNet10 dataset. It can be observed that the classification accuracy rapidly improves for data augmentation in the range of 1X to 3X. After that, there are still average improvements but not as significant.  
In contrast to prior work by Azizi et al.~\cite{azizi2023synthetic}, which reported performance degradation in ResNet-50 classification accuracy when the size of augmented data exceeded four times the original dataset. The results indicate that with \nameNoSpace, the performance of both ResNet-50 and ResNet-101 models increases with the scale of synthetic data augmentation. Specifically, we scale up to ten times the original size of the ImageNet10 dataset, which contains approximately $13,000$ images. As illustrated in Figure~\ref{fig:scaling_up}, validation accuracy trends upwards as the dataset size increases, without the performance degradation observed in the prior study. This suggests that \name does not compromise baseline performance, even at high augmentation scales.

\begin{figure*}[!ht]
\centering
\includegraphics[width=0.99\textwidth]
{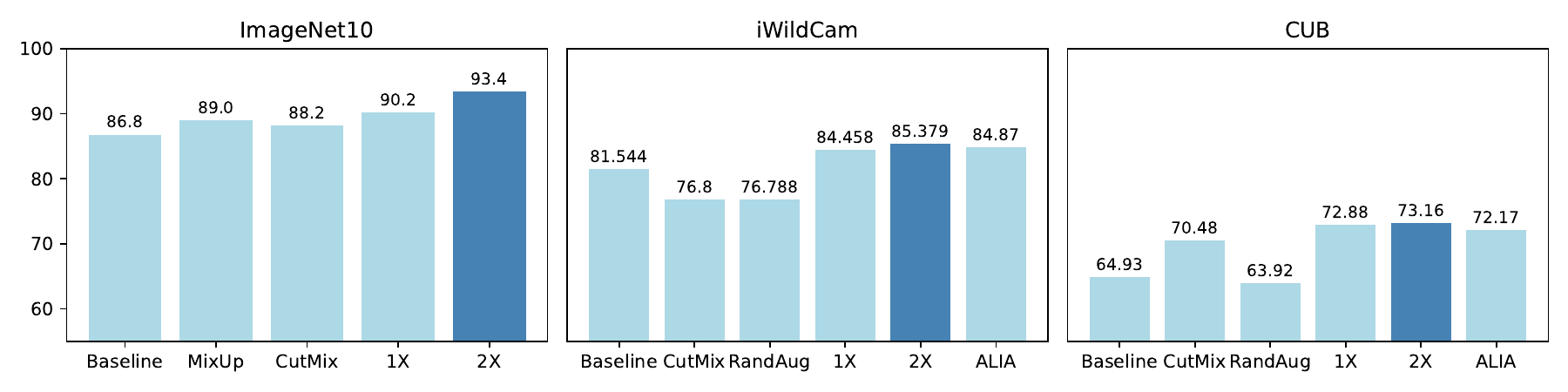}
\vspace{-6pt}
\caption{This figure presents bar charts comparing validation performance across the ImageNet10, iWildCam, and CUB datasets. For ImageNet10, the ResNet-50 model is trained from scratch with both original data and augmented data in various scales, shows AGA significantly outperforming other methods. In contrast, for iWildCam and CUB datasets, we employ a pretrained ResNet-50 model as used in ALIA. AGA consistently exceeds the performance of the baseline, CutMix, and RandAug at both one-time (1X) and two-time (2X) augmentation levels, and surpasses ALIA in the 2X augmentation scenario for iWildCam. For the CUB dataset, AGA again outperforms all competitors for both 1X and 2X augmentations.}
\label{fig:imagenet10_iwild_cub_results}
\vspace{-10pt}
\end{figure*}

\textbf{Comparison with SOTA:} We now turn our attention to comparing the performance of \name with previous approaches to data augmentation on the ImageNet10, iWild, and CUB data sets, which is shown in Figure~\ref{fig:imagenet10_iwild_cub_results}. The figure shows the performance of \name with data augmentation up to 3X, MixUP~\cite{zhang2017mixup}, CutMix~\cite{Yun_2019_ICCV}, RandAug~\cite{Cubuk_2020_CVPR_Workshops},and ALIA~\cite{dunlap2024diversify}. Recall that  former three methods are traditional data augmentation methods while the latter is based on generative AI. We only show results of RandAug and ALIA on iWild and CUB because the source code cannot easily be executed on ImageNet10 dataset.

Figure~\ref{fig:imagenet10_results} displays samples of images generated by \name for ImageNet10. The figure shows that \name successfully augments the input image with diverse backgrounds while preserving the properties of the foreground subject. We compare four ResNet models validation accuracy when the models are  trained with (1) real images and (2) augmented dataset at various scales. Figure~\ref{fig:imagenet10_iwild_cub_results} reports validation results of the respective models for \nameNoSpace, along with CutMix and MixUp on the ImageNet10 chart.  While CutMix, MixUp, and the baseline rely solely on original ImageNet10 images, our study extends to include augmented data up to two times the original dataset size (denoted as 1X to 2X). All models are evaluated using the same original ImageNet validation dataset. Our findings indicate that \name consistently outperforms both the baseline and other augmentation techniques across all tested scales.

Following this, we compare the performance of \name with other augmentation techniques such as CutMix, MixUp, and ALIA on the CUB and iWild datasets. Using \name, we generate synthetic images at various scales for these datasets and adhere to the same training methodology described in ALIA's scripts for a direct comparison. Figure~\ref{fig:imagenet10_iwild_cub_results} displays the validation performances on iWildCam, with our method surpassing all others at twice the augmentation scale.
 In addition to that, Figure~\ref{fig:imagenet10_iwild_cub_results} also depicts AGA outperforming competing approaches in CUB for both +1X and +2X augmentation scales.

\begin{table*}[htb]
\centering
\caption{Comparison of baseline and AGA-augmented training for four ResNet model variants (18, 50, 101, and 152) using the AIC, SIC, and insertion tests. We perform an in-distribution evaluation on ImageNet10-val and an out-of-distribution evaluation on ImageNet-V2 and ImageNet-Sketch datasets. We directly apply the trained models to these out-of-distribution evaluation datasets without further fine-tuning.}
\renewcommand{\arraystretch}{1.1} %
\small
\begin{tabular}{l|l|cc|cc|cc}
    \hline
    \multirow{3}{*}{\textbf{Metric}} & \multirow{3}{*}{\textbf{Model}} & \multicolumn{2}{|c}{\textbf{In-Distribution}} & \multicolumn{4}{|c}{\textbf{Out-of-Distribution}} \\
    \cline{3-8}
     &  & \multicolumn{2}{c}{\textbf{ImageNet10-Val}} & \multicolumn{2}{|c}{\textbf{ImageNet-V2}} & \multicolumn{2}{|c}{\textbf{ImageNet-S}} \\ 
    \cline{3-8}
     & & \textbf{Baseline} & \textbf{AGA} & \textbf{Baseline} & \textbf{AGA} & \textbf{Baseline} & \textbf{AGA} \\
     \hline
     \multirow{4}{*}{AIC($\uparrow$)}
     & ResNet-18 & $0.717$ & $\mathbf{0.764}$ & $0.694$ & $\mathbf{0.824}$ & $\mathbf{0.576}$ & $0.532$ \\
      & ResNet-50 & $0.792$ & $\mathbf{0.815}$ & $0.779$ & $\mathbf{0.799}$ & $\mathbf{0.670}$ & $0.539$ \\
      & ResNet-101 & $0.619$ & $\mathbf{0.843}$ & $0.682$ & $\mathbf{0.798}$ & $0.524$ & $\mathbf{0.598}$ \\
      & ResNet-152 & $0.599$ & $\mathbf{0.878}$ & $0.560$ & $\mathbf{0.865}$ & $0.521$ & $\mathbf{0.614}$ \\
      \hline
      \multirow{4}{*}{SIC($\uparrow$)}
      & ResNet-18 & $0.745$ & $\mathbf{0.816}$ & $0.811$ & $\mathbf{0.880}$ & $\mathbf{0.681}$ & $0.640$ \\
      & ResNet-50 & $0.816$ & $\mathbf{0.882}$ & $0.792$ & $\mathbf{0.877}$ & $\mathbf{0.679}$ & $0.639$ \\
      & ResNet-101 & $0.527$ & $\mathbf{0.866}$ & $0.579$ & $\mathbf{0.795}$ & $0.437$ & $\mathbf{0.654}$ \\
      & ResNet-152 & $0.557$ & $\mathbf{0.872}$ & $0.541$ & $\mathbf{0.860}$ & $0.488$ & $\mathbf{0.644}$ \\
      \hline
      \multirow{4}{*}{Insertion($\uparrow$)}
      & ResNet-18 & $0.220$ & $\mathbf{0.299}$ & $0.202$ & $\mathbf{0.271}$ & $0.173$ & $\mathbf{0.211}$ \\
      & ResNet-50 & $0.235$ & $\mathbf{0.320}$ & $0.215$ & $\mathbf{0.290}$ & $0.190$ & $\mathbf{0.248}$ \\
      & ResNet-101 & $0.115$ & $\mathbf{0.443}$ & $0.115$ & $\mathbf{0.386}$ & $0.120$ & $\mathbf{0.291}$ \\
      & ResNet-152 & $0.128$ & $\mathbf{0.446}$ & $0.120$ & $\mathbf{0.406}$ & $0.131$ & $\mathbf{0.294}$ \\
      \bottomrule
\end{tabular}
\label{tab:XAI}
\vspace{-5pt}
\end{table*}

\subsection{Evaluation of Impact on Generalizability}
Machine learning models typically struggle with out-of-distribution data, but models trained with AGA-augmented data show commendable performance in such cases. We assess our image augmentation method on ImageNet-Sketch and ImageNet-V2 datasets, training the CNN models with both original images and a combination of original and augmented images. For evaluation, we adhere to the same validation dataset across all models. The ImageNet-Val dataset is used for in-distribution testing, while validation data from ImageNet-Sketch and ImageNet-V2 are used for out-of-distribution testing.

Our results are summarized in Table~\ref{tab:in_out_comparison}, which includes performance metrics for four ResNet models on both in-distribution and out-of-distribution data, with the specific gains over baseline models quantified as $\Delta$ in the table. The table highlights up to $15.6$\% improvements in accuracy for the ResNet-101 model for {\bf ImageNet10-Val} when trained with AGA augmented data. We also see significant performance improvements for out-of-distributions, proving the fact that AGA augmented data increase generalizability of fine-grained classification models.

\begin{figure*}[htbp]
    \centering
    \begin{subfigure}[b]{0.49\textwidth} %
        \centering
        \includegraphics[width=3in, height=2.395in]{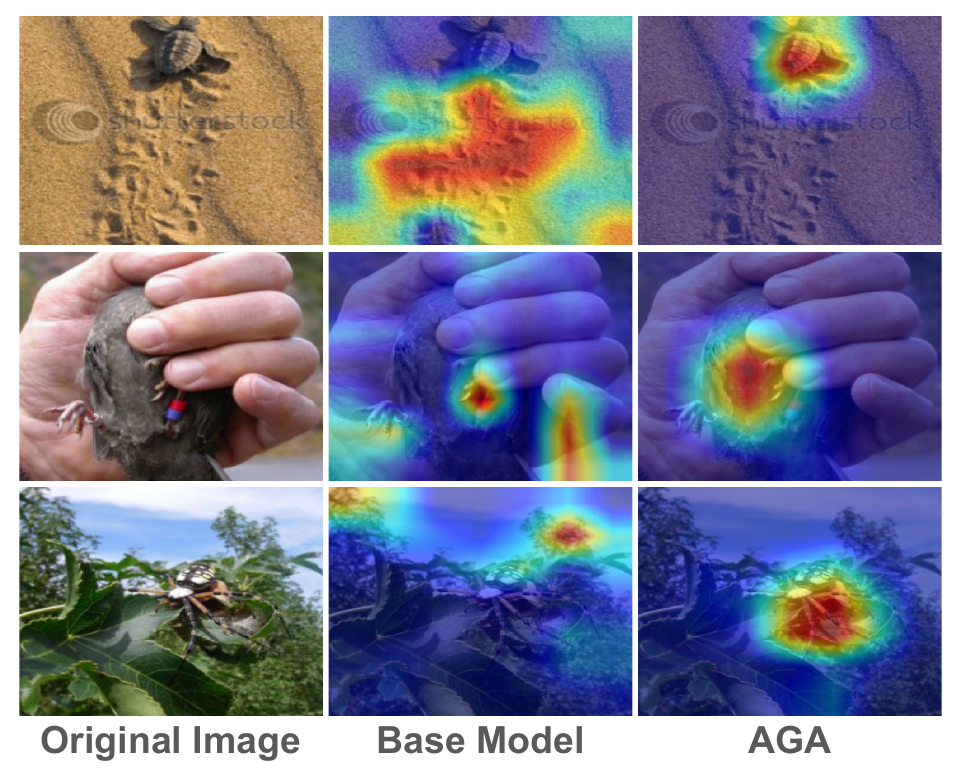}
        \caption{Baseline Model fails to identify samples correctly.}
        \label{fig:robust_image1}
    \end{subfigure}
    \hfill 
    \begin{subfigure}[b]{0.49\textwidth} %
        \centering
        \includegraphics[width=3in, height=2.395in]{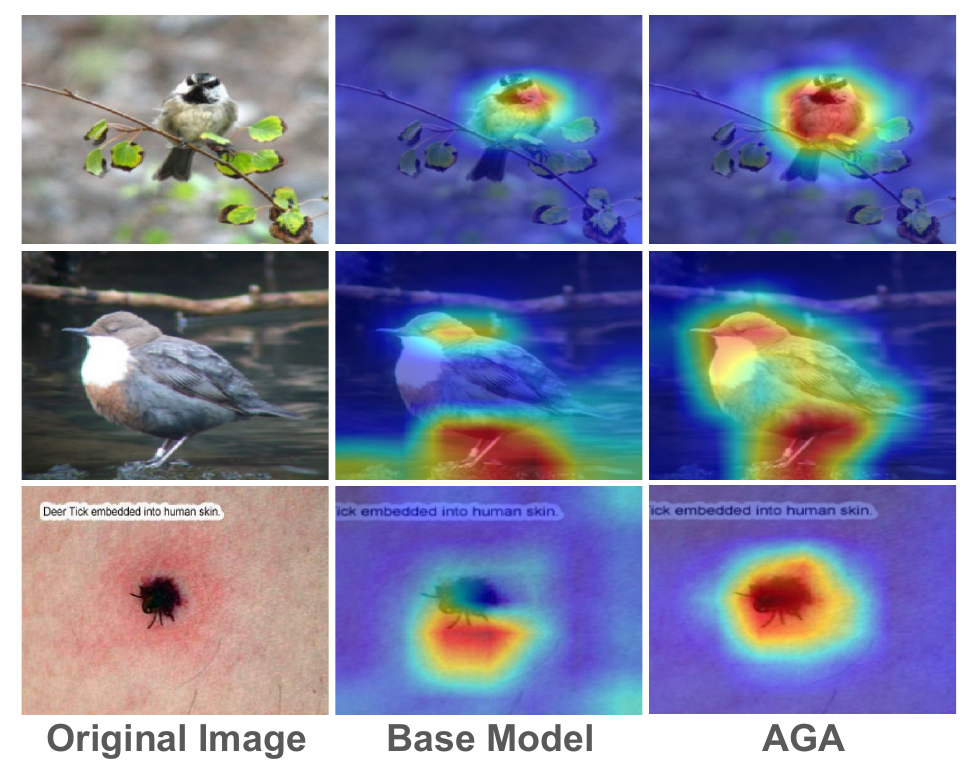}
        \caption{Baseline \& augmented models  are correct.}
        \label{fig:robust_image2}
    \end{subfigure}
    \caption{ This figure shows the impact of the data augmentation on explainability using feature attributions computed using GradCam~\cite{Selvaraju_2017_ICCV}. (a) images only correctly classified by the classifier trained using data augmentation. (b) images correctly classified by both the original model and the model trained with data augmentation. The model trained with only the original real data fails to identify the bird correctly and focuses on the scatter pixel region. However, it can be observed that even when both models provide the correct classification, the augmented model provides better attributions of the object. More visualization results are presented in supplementary material.}
    \label{fig:robustness}
\end{figure*}

\subsection{Evaluation of Impact on  Explainability}

Explainability is an increasingly critical aspect of AI, particularly in understanding how machine learning models make decisions. Our study explores the impact of subject-oriented data augmentation provided by AGA on model explainability. By enhancing image diversity through augmentation, we aim to develop more robust and interpretable classifiers. We train models on both the baseline ImageNet10 dataset and augmented data to compare performance. For visualizing how models focus on relevant areas within images, we employ GradCam~\cite{Selvaraju_2017_ICCV}, a tool that highlights significant regions influencing model decisions. In our findings, as shown in Figure~\ref{fig:robustness}, we compare models at the $85^{th}$ epoch, trained solely on real ImageNet10 data and those trained on ImageNet10 augmented data. The model trained only on real data incorrectly classifies three specific images (Figure~\ref{fig:robust_image1}), whereas the model trained with AGA-augmented data correctly identifies these images. GradCam visualizations reveal that the baseline model often focuses on irrelevant pixels, whereas the AGA-trained model more accurately targets pixels within the subject area. This explains that the augmented data helps the model to learn correctly. Further comparisons using images correctly classified by both models (Figure~\ref{fig:robust_image2}) show that the AGA-augmented model more consistently identifies correct subject areas, underscoring the benefits of diverse training data for improved model accuracy and explainability.

Additionally, we conduct quantitative explainability analysis utilizing performance information curves (PICs)~\cite{9008576}, which include two components: the softmax information curve (SIC) and the accuracy information curve (AIC). 
The PICs serve as a metric to assess model performance relative to the informational content (entropy) present in the input data. 
The SIC reflects the softmax value for the input's original class, contributes to the model explainability assessment. Moreover, an insertion test~\cite{Petsiuk2018rise} was also conducted to gauge model training performance across different methods. Table~\ref{tab:XAI} presents the AIC, SIC, and insertion test outcomes for various ResNet model variants (18, 50, 101, and 152). It was observed that the AGA-augmented models generally exhibited improved performance across most cases, barring two instances in both AIC and SIC evaluations. This discrepancy can be attributed to the out-of-distribution nature of the ImageNet-V2 and ImageNet-Sketch datasets relative to the models trained on the ImageNet10 dataset and its augmented variant. The augmentation with additional data samples led to increased confusion in smaller models when determining the correct class, impacting their performance in certain scenarios.

\section{Conclusion and Future Work}
\label{sec:conclusion}

We introduce \nameNoSpace, a novel data augmentation method designed to address data scarcity in fine-grained image recognition. Our approach integrates image segmentation, automated background caption generation, and diffusion-based image synthesis to diversify backgrounds while maintaining the subject's integrity, thus enhancing training datasets for improved fine-grained classification performance, especially in low-data situations. \name reveals that additional generated data assists the deep learning model in concentrating on the expected subject regions, as evidenced by the Grad-CAM attribution method. The framework also demonstrates strong generalization on out-of-distribution data.
\name experiences compatibility issues concerning proper subjects and backgrounds, and occasionally produces visually inconsistent synthetic images by combining subjects with inappropriate backgrounds. This limitation underscores the potential for future research to explore new methods for generating images that maintain subject integrity while ensuring compatibility with backgrounds.

\bibliographystyle{unsrt}
\bibliography{egbib}

\appendix 
\section{Appendix}
In this appendix, we provide supplementary technical details and experiments that could not fit within the main manuscript. We present detailed information about all reference datasets used, encompassing training and validation samples, in Section~\ref{sec:additional_dataset_details}. All the information about the CNN model training procedure, as well as details about all the hyperparameters used during training and validation, is shown in Section~\ref{sec:Training_and_Hyperparameter}. Lastly, we present qualitative visualization results, encompassing synthetic images generated by the \name, GradCam visualization heatmaps for enhanced explainability, and UMAP plots depicting feature clusters to assess the quality of generated image features in Section~\ref{sec:additional_results}.

\subsection{Additional Dataset Details}
\label{sec:additional_dataset_details}

In this section, we present additional details about all the representative datasets we used to evaluate our proposed method \name. We use the ImageNet10 dataset, which is a subset of the original ImageNet dataset~\cite{deng2009imagenet} with $10$ different classes. These are chickadee (\textit{n01592084}), water ouzel (\textit{n01601694}), loggerhead (\textit{n01664065}), box turtle (\textit{n01669191}), garter snake (\textit{n01735189}), sea snake (\textit{n01751748}), black and gold garden spider (\textit{n01773157}), tick (\textit{n01776313}), ptarmigan (\textit{n01796340}), prairie chicken (\textit{n01798484}). We use the training and validation sets from ImageNet~\cite{deng2009imagenet} for these 10 classes. We also utilize the iWildCam~\cite{koh2021wilds} dataset, which contains a large collection of global camera trap images of $7$ different classes of background, elephant, impala, cattle, zebra, dik-dik, and giraffe, and the CUB~\cite{wah2011caltech} dataset, a fine-grained classification set of $200$ bird species from Flickr. We maintain the same data distribution ratio as in the previous work~\cite{dunlap2024diversify} for the train and test sets to ensure a fair comparison. To show the robustness and generalization capability of our method, we additionally use two other datasets named ImageNet-Sketch~\cite{wang2019learning} and ImageNet-V2~\cite {recht2019imagenet}, where ImageNet-Sketch is the sketch version and ImageNet-V2 is the reproduced version of ImageNet. We utilize ImageNet-Sketch and ImageNet-V2 to validate the robustness of \name against out-of-distribution samples. The number of training and validation images used for evaluation is presented in Table~\ref{tab:dataset_table}.

\begin{table}[!ht]
\centering
\caption{Number of train and validation samples for all the representative datasets utilized in \textit{\name}. We use '-' to denote the absence of training samples on the ImageNet-Sketch and ImageNet-V2 datasets. These datasets are used for out-of-distribution validation to assess the robustness of the \name framework.}
\renewcommand{\arraystretch}{1.2} %
\small
\begin{tabular}{l|cc}
    \hline
    \multirow{2}{*}{\textbf{Dataset Name}} & \multicolumn{2}{c}{\textbf{No. of Images}} \\
     & \textbf{Training}  & \textbf{Validation}   \\
    \hline 
    ImageNet10~\cite{deng2009imagenet} & 13046 & 500 \\
    iWildCam~\cite{koh2021wilds} & 6052 & 8483 \\
    CUB~\cite{wah2011caltech} & 4994 & 5794 \\
    ImageNet-Sketch~\cite{wang2019learning} & - & 511 \\
    ImageNet-V2~\cite{recht2019imagenet} & - & 102 \\
    \hline
\end{tabular}
\label{tab:dataset_table}
\end{table}

\begin{figure*}[ht]
    \centering
    \begin{subfigure}[b]{0.32\textwidth}
        \centering
        \includegraphics[width=\textwidth]{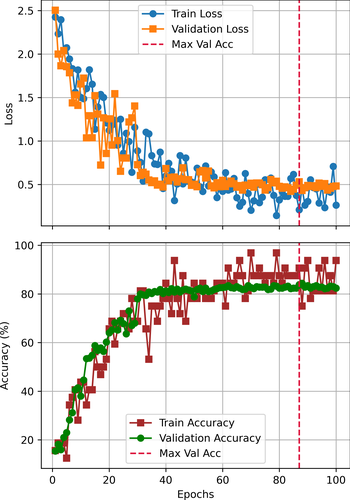}
        \caption{Base model training}
        \label{fig:train_and_test_base}
    \end{subfigure}
    \hfil
    \begin{subfigure}[b]{0.32\textwidth}
        \centering
        \includegraphics[width=\textwidth]{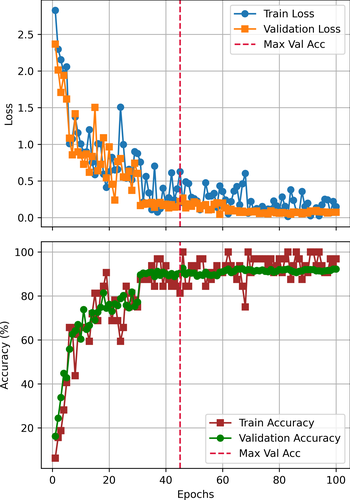}
        \caption{AGA model training}
        \label{fig:train_and_test_aga}
    \end{subfigure}
    \caption{Training loss, validation loss, and accuracy curve for the base model for real data samples and the AGA model for augmented datasamples of ImageNet 10. The training and validation losses exhibit downward trends for both the base and AGA model training, with the validation loss remaining relatively stable and not showing a significant upward trend. This indicates that the models are generalizing well to unseen data and do not show signs of overfitting.}
    \label{fig:train_and_test}
\end{figure*}

\subsection{Training and Hyperparameter Details}
\label{sec:Training_and_Hyperparameter}
Our automatic image augmentation framework \name starts by separating the main subjects in the image using segmentation methods. Then, it uses a large language model (LLM) to generate different captions of backgrounds. These captions are fed into a vision model like Stable Diffusion to create various backgrounds. In the end, \name combines the separated subjects with the newly created backgrounds. We utilize a Llama-2-13B-GPTQ from Hugging-Face~\cite{von-platen-etal-2022-diffusers} to create background image captions and Stable Diffusion XL~\cite{podell2023sdxl} text-to-image model to generate background image, with default hyperparameters.

After the generation of augmented images we evaluate the quality of the additional data samples using several CNN classifier models. We employ ResNet variants {18, 50, 101, 152} as the classification models for training. We train these CNN models from scratch using PyTorch's standard training script~\cite{torchvision2016} which includes PyTorch's default hyperparameter set~\cite{pytorchTrainStateOfTheArt}.
All the hyperparameter values used for CNN classifier training are presented in Table~\ref{tab:hyper_table}. We train all the classifier models multiple times and report the average performance. While training these CNN models, we carefully addressed the issue of overfitting. We often refer to the maximum classifier accuracy for any epoch by avoiding overfitting. We ensure this by using the training and validation loss. We train all the models in such a way that the difference between the training and validation loss is minimized. The training and validation losses of ImageNet10 training are presented in Figure~\ref{fig:train_and_test}, with real data shown in~\ref{fig:train_and_test_base} and augmented data in~\ref{fig:train_and_test_aga}. The x-axis represents the number of epochs, and the y-axis represents both accuracy and loss values. In both cases, the red vertical dashed line represents the epoch at which we achieve the maximum validation accuracy. We observe that both training and validation losses exhibit downward trends in Figure~\ref{fig:train_and_test} for both base and AGA model training. No such scenario is detected where training loss keeps decreasing while validation loss starts to increase. The validation loss remains relatively stable and doesn't show a significant upward trend. This indicates the model is generalizing well to unseen data and clearly shows no signs of overfitting.

\begin{table}[!ht]
\centering
\caption{Training Details of ResNet Models}
\renewcommand{\arraystretch}{1.2} %
\small
\begin{tabular}{l|l}
    \hline
    \textbf{Model Parameter} & \textbf{ResNet-\{18,50,101,152\}}  \\
    \hline 
    Epochs & 100 \\
    Batch Size & 32  \\
    Optimizer & Stochastic Gradient Descent (SGD) \\
    Momentum & 0.9 \\
    Learning Rate & 0.1 \\
    Learning Rate Scheduler & StepLR  \\
    Learning Step Size & 30 \\
    Gamma Parameter & 0.1 \\
    Weight Decay & 1e-4 \\
    Interpolation & Bilinear \\
    Loss Function & CrossEntropyLoss \\
    \hline
\end{tabular}
\label{tab:hyper_table}
\end{table}

\subsection{Additional Results}
\label{sec:additional_results}

We present additional synthetic images generated by \name in Section~\ref{sec:syn_image}. We also exhibit more GradCam visualization results to demonstrate the improved explainability of the classifier model trained with augmented data samples compared to one trained with only real samples in Section~\ref{sec:vis_res}. Moreover, we display the CNN model-extracted features in a UMAP plot, showing feature clusters for different classes of real and augmented images in Section~\ref{sec:feature_cluster}.

\subsubsection{Additional Synthetic Images}
\label{sec:syn_image}
\begin{figure*}[!h]
\centering
\includegraphics[width=0.78\textwidth]{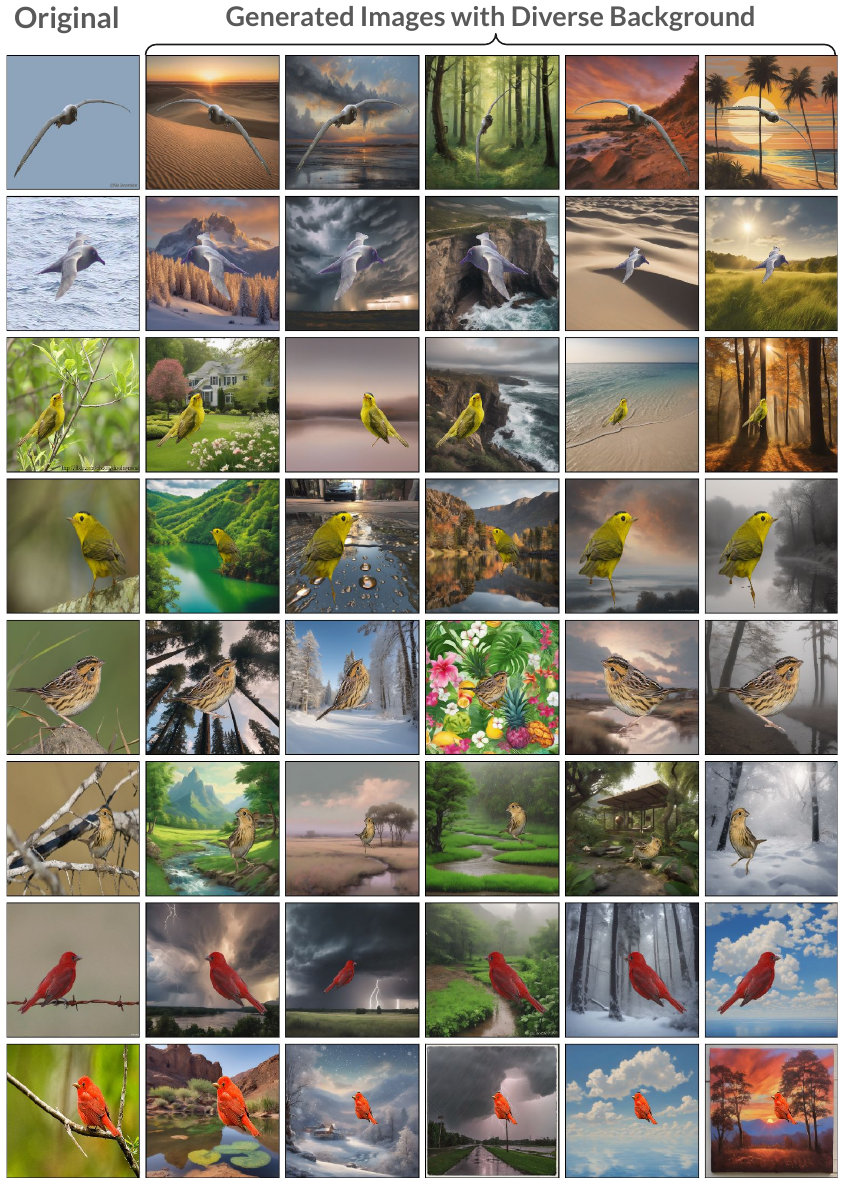}
\caption{The figure displays the original image samples from CUB and the generated images using \name.}
\label{fig:synthetic_1}
\end{figure*}

\begin{figure*}[!h]
\centering
\includegraphics[width=0.78\textwidth]{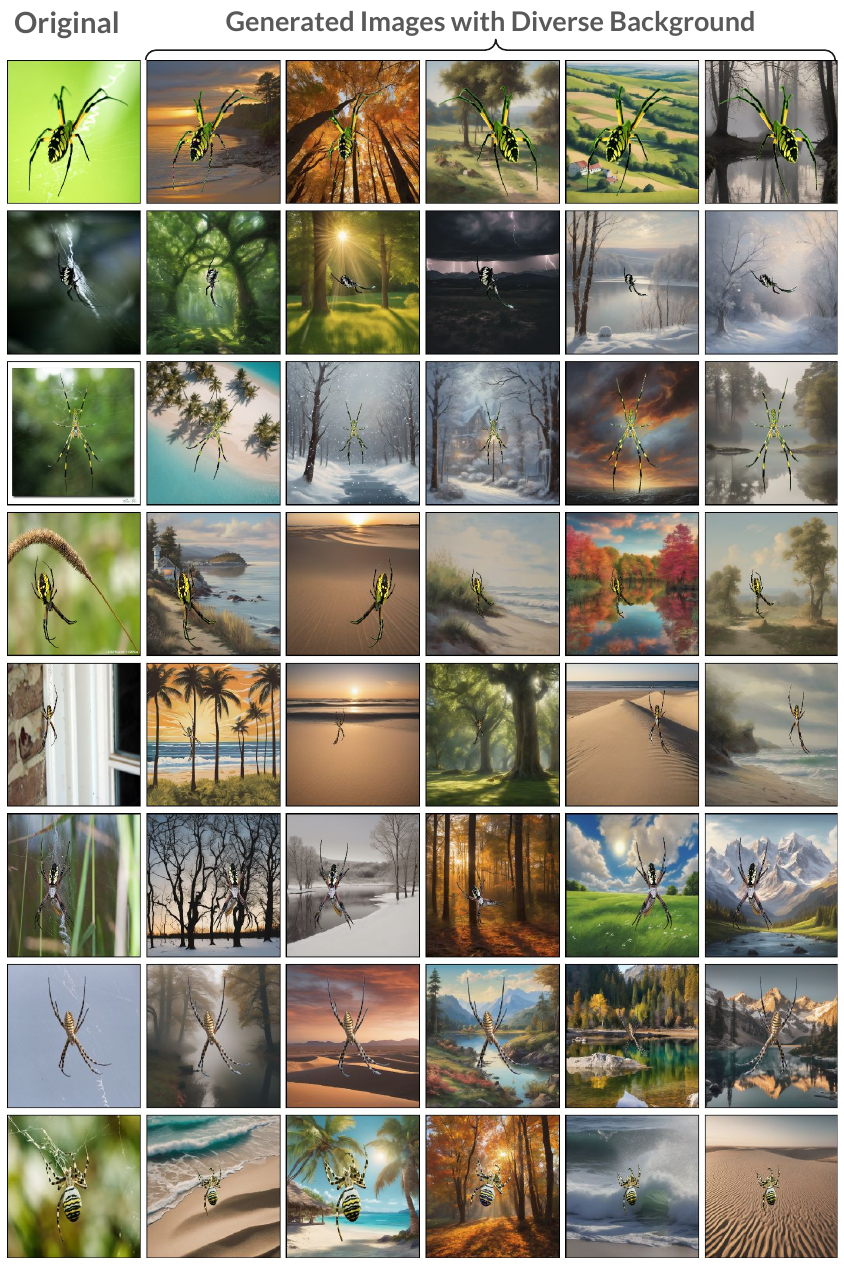}
\caption{The figure displays the original image samples from ImageNet10 and the generated images using \name.}
\label{fig:synthetic_2}
\end{figure*}

\begin{figure*}[!h]
\centering
\includegraphics[width=0.78\textwidth]{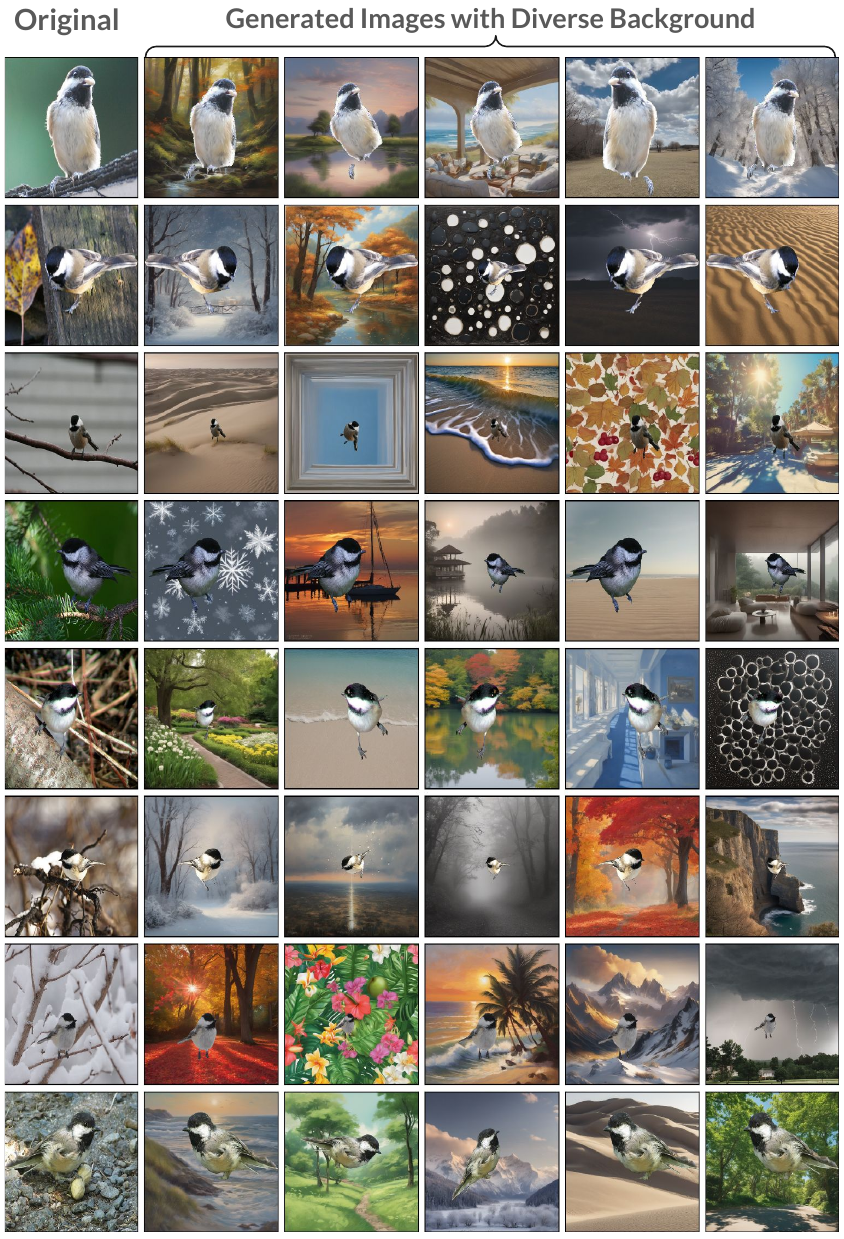}
\caption{The figure displays the original image samples from ImageNet10 and the generated images using \name. }
\label{fig:synthetic_3}
\end{figure*}

We present more generated images from real image with diverse backgrounds. Figure~\ref{fig:synthetic_1}, \ref{fig:synthetic_2} and \ref{fig:synthetic_3} display multiples synthetic images generated by \name using ImageNet10 and CUB traing image samples.

\clearpage

\clearpage

\clearpage

\subsubsection{GradCam Visualization Results}
\label{sec:vis_res}
We present additional GradCam visualization results here to show the explanable capability of the base model and AGA model for ImageNet 10. The base model classifier is trained with only the real images of the ImageNet10 dataset, but the AGA model is trained with real images as well as augmented images generated by the AGA method.

\begin{figure}[!h]
    \hspace*{40pt}
    \begin{minipage}{0.85\textwidth}
        \centering
        \includegraphics[width=\linewidth]{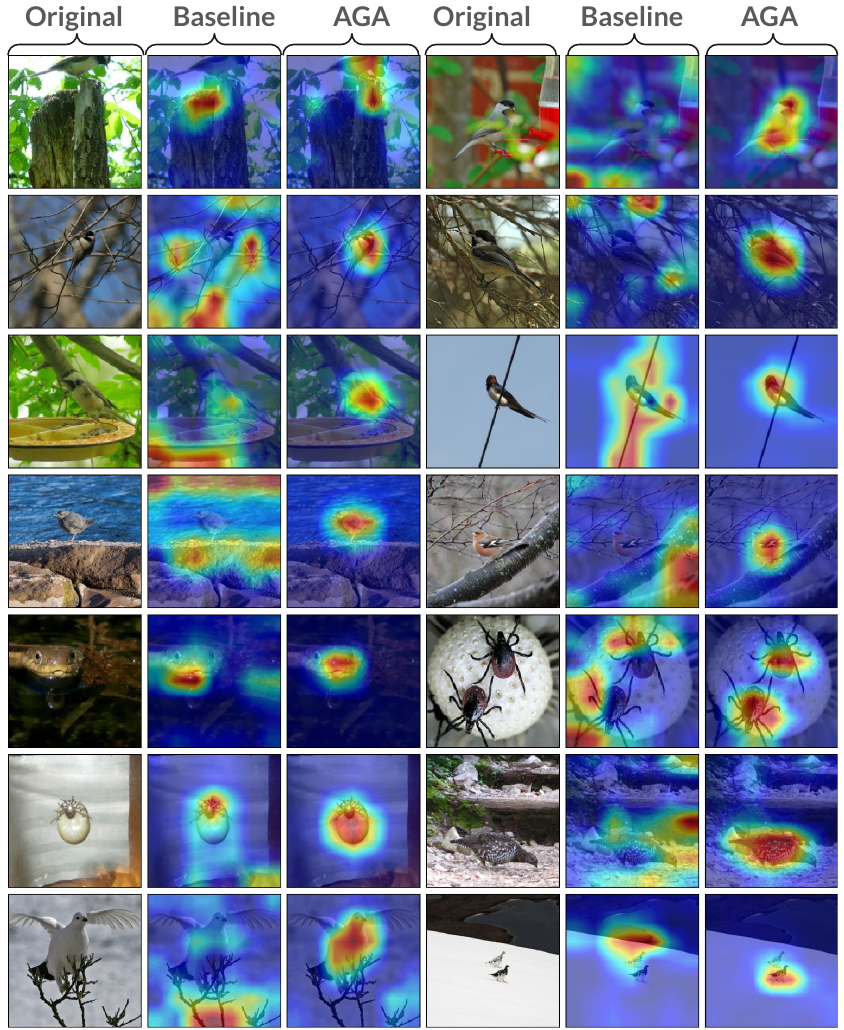}
        \caption{The figure displays additional GradCam visualization results for ImageNet10 dataset.}
        \label{fig:gradcam}
    \end{minipage}
\end{figure}

We demonstrate several validation dataset samples of ImageNet10 that are misclassified by base model in Figure~\ref{fig:gradcam}. On the other hand, the AGA model correctly classified these data samples, and the following GradCam visualizations reveal that the baseline model often focuses on irrelevant pixels, whereas the AGA-trained model more accurately targets pixels within the subject area.

\clearpage

\subsubsection{Feature Cluster}
\label{sec:feature_cluster}

\begin{figure}[h]%
    \vspace{-20pt}
    \centering
    \subfloat[\centering Features extracted from ImageNet10 real images]{{\includegraphics[width=9cm]{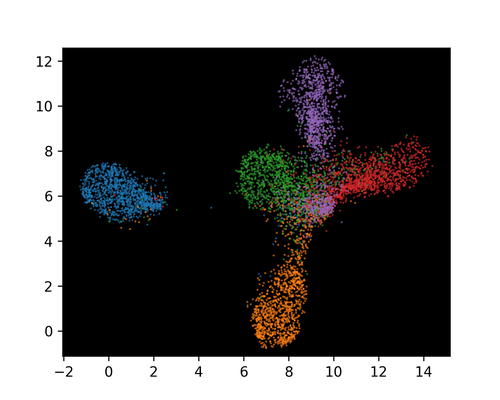} }}%
    \qquad %
    \subfloat[\centering Features extracted from augmented images generated by \name]{{\includegraphics[width=9cm]{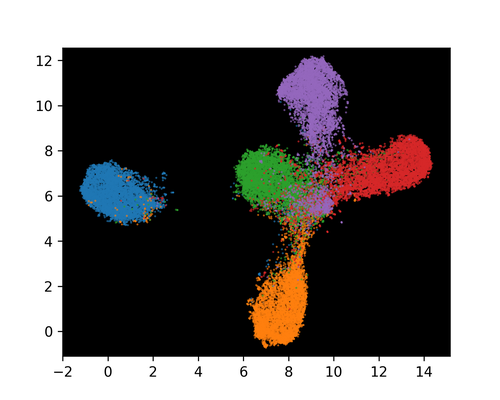} }}%
    \caption{UMAP plot of feature clusters of five distinct classes of ImageNet10 dataset where features are extracted from last layer of ResNet-50 model. }%
    \label{fig:feature_cluster}%
\end{figure}

We conducted an additional experiment to verify that additional synthetic images do not introduce irrelevant features. We utilize the last-layer feature outputs from the ResNet-50 model for both ImageNet10 real and AGA-augmented images. Each image yields 2048 features, which we use to plot feature clusters. We illustrate five distinct class clusters of ImageNet10 for both real and AGA-augmented images in Figure~\ref{fig:feature_cluster}. The figure shows that additional generated images enhance cluster density without significantly increasing inter-cluster distances.

\end{document}